%% file: main.tex
\documentclass[10pt,twocolumn,letterpaper]{article}

\usepackage{iccv}
\usepackage{times}
\usepackage{epsfig}
\usepackage{graphicx}
\usepackage{amsmath}
\usepackage{amsfonts,,amssymb}
\usepackage{booktabs}

\usepackage{color}
\usepackage{bm,xspace}
\usepackage{comment}
\usepackage{bbold}
\usepackage{verbatim}
\usepackage{balance}
\usepackage{etoolbox,siunitx}
\usepackage{calc}
\usepackage{pifont,hologo}
\usepackage{adjustbox}
\usepackage{tabularx}
\usepackage{multirow}
\usepackage{caption} 
\usepackage{overpic}
\usepackage{array,makecell} 
\usepackage{esvect} 
\usepackage{textcomp, gensymb} 
\usepackage{tablefootnote}
\usepackage{arydshln}
\usepackage{amsmath}
\usepackage[skip=3pt]{subcaption}

\usepackage{algorithm2e}

\definecolor{mygray}{gray}{0.9} 
\definecolor{myskyblue}{RGB}{230, 249,255}
\definecolor{myyellowgreen}{RGB}{240, 240, 245}

\newcolumntype{?}[1]{!{\vrule width #1}} 

\newcolumntype{Y}{>{\centering\arraybackslash}X}

\usepackage[pagebackref=true,breaklinks=true,letterpaper=true,colorlinks,bookmarks=false]{hyperref}

\usepackage[capitalize]{cleveref}
\crefname{section}{Sec.}{Secs.}
\Crefname{section}{Section}{Sections}
\Crefname{table}{Table}{Tables}
\crefname{table}{Tab.}{Tabs.}

\iccvfinalcopy 

\def\iccvPaperID{3642} 

\ificcvfinal\fi

\begin{document}

\title{Dynamic Voxel Grid Optimization for High-Fidelity RGB-D Supervised \linebreak Surface Reconstruction}

\author{Xiangyu Xu\textsuperscript{1}\thanks{Joint first authorship. L. Chen's contribution was made while interning at OPPO US Research Center.} \and Lichang Chen\textsuperscript{2,1}\footnotemark[1] \and Changjiang Cai\textsuperscript{1} \and Huangying Zhan\textsuperscript{1}
\and Qingan Yan\textsuperscript{1}\thanks{Corresponding author (yanqinganssg@gmail.com)} \and Pan Ji\textsuperscript{1} \and Junsong Yuan\textsuperscript{3} \and Heng Huang\textsuperscript{2} \and Yi Xu\textsuperscript{1} \and
\textsuperscript{1}OPPO US Research Center, InnoPeak Technology, Inc \\
\textsuperscript{2}University of Maryland \\
\textsuperscript{3}State University of New York at Buffalo
}


\maketitle
\ificcvfinal\thispagestyle{empty}\fi

\begin{abstract}
\input{abstract}
\end{abstract}

\section{Introduction}
\label{sec:intro}
\input{intro}

\section{Related Work}\label{sec:related}
\input{related}

\section{Method}\label{sec:method}

\input{method}

\section{Experiments}\label{sec:experiments}
\input{experiments}

\section{Conclusions}\label{sec:conclusions}
\input{conclusions}

\section*{Appendix}
\input{appendix}


{\small
\bibliographystyle{ieee_fullname}
\bibliography{my_ref}
}

\end{document}

%% file: abstract.tex
Direct optimization of interpolated features on multi-resolution voxel grids has emerged as a more efficient alternative to MLP-like modules. However, this approach is constrained by higher memory expenses and limited representation capabilities. In this paper, we introduce a novel dynamic grid optimization method for high-fidelity 3D surface reconstruction that incorporates both RGB and depth observations. Rather than treating each voxel equally, we optimize the process by dynamically modifying the grid and assigning more finer-scale voxels to regions with higher complexity, allowing us to capture more intricate details. Furthermore, we develop a scheme to quantify the dynamic subdivision of voxel grid during optimization without requiring any priors. The proposed approach is able to generate  high-quality 3D reconstructions with fine details on both synthetic and real-world data, while maintaining computational efficiency, which is substantially faster than the baseline method NeuralRGBD~\cite{azinovic2022neural}.

%% file: intro.tex
3D scene reconstruction plays an essential role in a variety of virtual reality (VR), augmented reality (AR), and robotics applications. Although low-cost depth sensors have proven to be an efficient tool for capturing 3D geometries, relying exclusively on range measurements, such as in the case of KinectFusion~\cite{newcombe2011kinectfusion}, may result in incomplete or incorrect reconstructed geometry due to deficiencies in depth data. To address this issue, radiance-based scene representations~\cite{azinovic2022neural,wang2022go} offer a powerful way to incorporate spatial context from color observations for more high-quality RGB-D reconstruction.

\begin{figure}[t!]
    \centering
	\includegraphics[width=1.0\columnwidth]{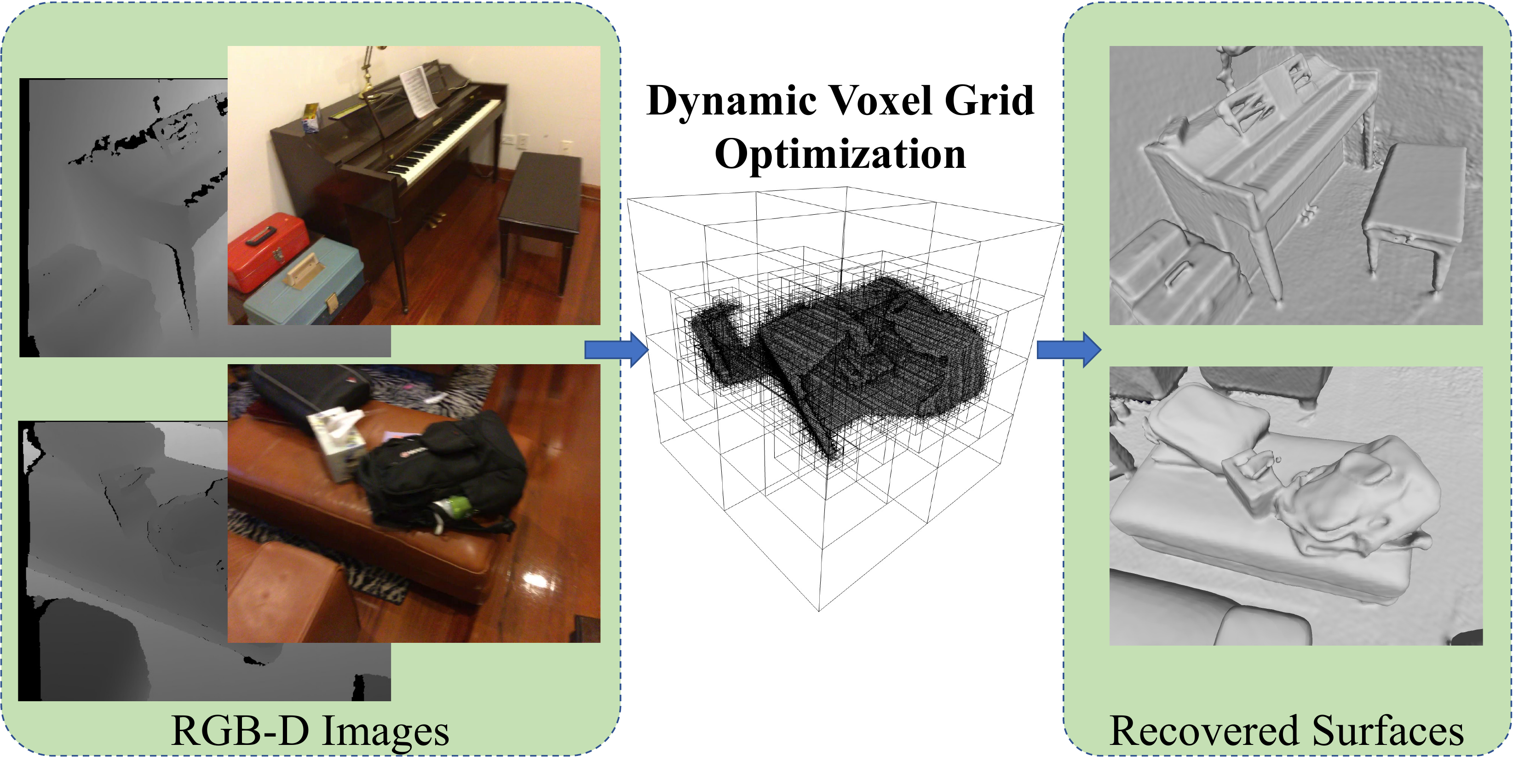}
	\caption{Our proposed dynamic voxel grid optimization method can generate high-fidelity 3D reconstructions from RGB-D data. The core idea is to dynamically refine the grid representation during optimization by assigning more finer-scale voxels to regions with higher complexity, and efficiently regress 3D structures via interpolation rather than MLPs.} 
	\label{fig:teaser}
\end{figure}

%

Drawing from~\cite{mildenhall2021nerf}, NeuralRGBD~\cite{azinovic2022neural} employs two multi-layer perceptrons (MLP) to learn the mapping from a 3D query to its corresponding signed distance function (SDF) and color value. Despite the effectiveness of scene reconstruction, this comes at the cost of hours of training time. While Go-Surf~\cite{wang2022go} cuts the runtime into minutes, it can not well preserve high-frequency details since the limited multi-scale resolution.
Taking inspiration from the recent success~\cite{fridovich2022plenoxels,sun2022direct} that uses voxel grid to explicitly store rendering properties, we introduce a dynamic voxel grid optimization for RGB-D supervised surface reconstruction. Our approach seeks to dynamically partitions each voxel in optimization according to local scene complexity. In this way, regions with more geometric and textural information would be represented by voxels of higher resolution, while surfaces with simple details remain in relatively low resolution. Divergent from Octree structure~\cite{yu2021plenoctrees} which relies on a pretrained geometric prior, our scheme has to adaptively select certain voxels for subdivision during optimization in the absence of any known priors. To address this challenge, we measure depth and color inconsistency with the corresponding references as a means of quantification. It is intuitive that complex structures and textures with higher losses usually deserve more primitive samples to approximate. 

The proposed hierarchical representation has three merits. First, it helps improve scalability. While voxel-based architectures~\cite{wang2022go} lead to super-efficient convergences, such structures entail high memory consumption as the size of the scene grows. In contrast, our representation achieves better memory efficiency thanks to selective partitioning. Second, the representation has an inherent advantage in recovering geometric details. Specifically, with lower memory cost, our method is able to divide complex regions into voxels with higher resolution under the same hardware configuration, which accordingly alleviates suboptimal solutions and results in finer reconstructions. Moreover, unlike previous radiance-based 3D reconstruction methods~\cite{azinovic2022neural,wang2022go}, the proposed representation can be optimized simply via gradient descents and regularization without any MLPs and pre-training. Evaluations on both synthetic and real-world datasets show that our proposed method achieves similar runtime compared to concurrent work~\cite{wang2022go}. Meanwhile, our method produces 3D reconstruction with finer geometric details, as shown in \cref{fig:teaser}.

To summarize, our contributions are the following:

\begin{itemize}
    \item A novel 3D surface reconstruction method that directly regresses SDF from the supervision of calibrated color and depth images without leveraging any MLP components;
    \item A hierarchical structure that enables highly efficient scene representation and detailed geometry recovery;
    \item A well-designed partitioning strategy for dynamic voxel subdivision during optimization.
\end{itemize}

%% file: related.tex
Our work aims at reconstructing 3D objects from a sequence of calibrated RGB-D images. In addition to reviewing the representatives of classic RGB-D fusion methods and concurrent neural RGB-D reconstruction works, we also review the recent progress of NeRF and direct voxel grid optimization.

\subsection{Classical RGB-D Fusion}
There is a wide range of landmark 3D reconstruction methods purely relying on explicit geometric constraints from ordinary images without any implicit neural inference, such as Shape-from-Silhouette~\cite{laurentini1994visual}, Structure-from-Motion (SfM)~\cite{schonberger2016structure,snavely2006photo,yan2017distinguishing}, and SLAM~\cite{ji2022georefine,mur2015orb,qin2018vins}. The advancement of consumer depth cameras promotes research interest on RGB-D reconstruction. Methods such as~\cite{dai2017bundlefusion,newcombe2011kinectfusion} achieve great performance in reconstructing dense room-scale scenes and are mostly built upon a signed distance function (SDF)~\cite{curless1996volumetric}. KinectFusion~\cite{newcombe2011kinectfusion} provides an online depth fusion system to reconstruct indoor objects in real-time via a handheld RGB-D camera. To mitigate accumulated errors in long-range scanning, BundleFusion~\cite{dai2017bundlefusion} draws upon keypoint observations across color views to refine camera poses.~\cite{niessner2013real} proposes an interesting voxel hashing strategy for efficient volumetric reconstruction in large environments.~\cite{newcombe2015dynamicfusion} uses RGB-D images to reconstruct dynamic objects. Despite the effectiveness in real-time reconstruction, their performance relies highly on the accuracy of input range data and would be easily affected by physical sensor deficiencies.~\cite{fu2020joint,maier2017intrinsic3d} leverage the joint optimization for both texture and geometry. However the improvement is limited especially for complex scene surfaces.

\subsection{Neural Radiance Field}
Unlike traditional discretized volumetric representations, NeRF~\cite{mildenhall2021nerf} utilizes MLPs to map 3D coordinates into density and RGB color, which achieves remarkable quality in novel view synthesis. However, NeRF models require substantial computations in both training and rendering. Instead of applying deep MLPs, TensorRF~\cite{chen2022tensorf} models the scene with a 3D voxel grid and per-voxel features. It further factorizes the 4D tensors into low-rank components to accelerate the overall training. By using multi-resolution hashing, Instant-NGP~\cite{muller2022instant} enables efficient encoding and achieves remarkable speedup during training.
\par Besides accelerating the training side, there has also been plenty of interest in speeding up rendering. NSVF~\cite{liu2020neural} employs the octree and free space skipping to accelerate rendering. By utilizing a series of tiny MLPs instead of deep MLPs, KiloNeRF~\cite{reiser2021kilonerf} makes real-time rendering practical. DoNeRF~\cite{neff2021donerf} restricts the volume rendering and sampling to the near-surface area and achieves a significant decrease in inference cost.


%
\begin{figure*}[ht]
    \centering
	\includegraphics[width=0.9\textwidth]{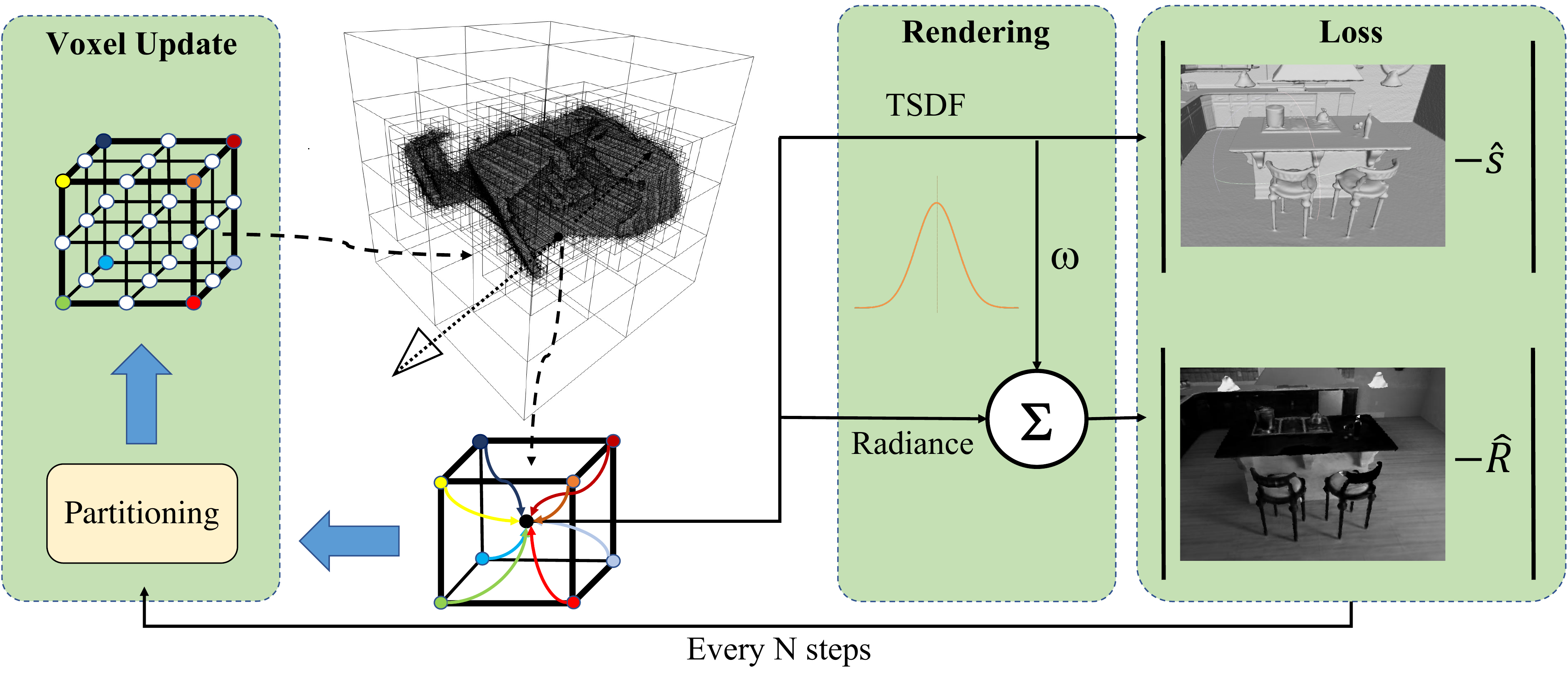}
	\caption{Our proposed pipeline utilizes a single dynamic voxel grid to efficiently store both Signed Distance Function (SDF) and color values. At each sample point along a ray, we trace the terminal voxel and compute the corresponding feature using trilinear interpolation of the voxel's eight corners. Loss terms are then applied to both the SDF values and color, and the voxel grid is updated based on the corresponding loss values after a specified number of steps.} 
	\label{fig:method}
\end{figure*}

\subsection{Neural RGB-D Surface Reconstruction}
Neural surface reconstruction~\cite{azinovic2022neural,li2022bnv,niemeyer2020differentiable,sun2021neuralrecon,wang2022go,wang2021neus,yariv2021volume,yariv2020multiview}, which reduces artifacts from traditional 3D reconstruction through training end-to-end networks, has been a trend for high-fidelity 3D surface reconstruction. In particular, for RGB-D reconstruction, NeuralRGBD~\cite{azinovic2022neural} first couples the merit of SDF representation and volume rendering to jointly leverage the context in color and depth. While it is able to reconstruct high-quality metrical 3D geometries, due to the similar network architecture to NeRF, NeuralRGBD takes hours of time to converge. GO-Surf~\cite{wang2022go} proposes a learned hierarchical feature voxel grid to represent the scene and achieves a substantial speedup over NeuralRGBD. Since it favors scene completeness, so the reconstructed result is prone to lose some high-frequency details. NiceSLAM~\cite{zhu2022nice} introduces a real-time dense RGB-D SLAM system that optimizes a multi-level feature grid representation with pre-trained geometric priors. NicerSLAM~\cite{zhu2023nicer} alleviates the requirement for depth images. Instead, it is enhanced with an additional monocular depth estimation module and a normal calculation module. 

\subsection{Direct Voxel Grid Optimization}
Directly optimizing interpolated features recorded in a voxel grid leads to orders of magnitude faster training and inference time than querying MLPs. To accelerate the training of NeRF models, DirectVoxGO~\cite{sun2022direct, sun2022improved} methods use a dense voxel grid to directly model volume density and optimize view-dependent colors in a hybrid way by incorporating voxel grid with shallow MLP. PlenOctrees ~\cite{yu2021plenoctrees} adopts an Octree structure to exclude redundant queries, however, an additional conversion step from a pretrained model is required to serve as the prior for structure generation. Plenoxel~\cite{fridovich2022plenoxels} also utilizes the Octree structure to model the scene with additional spherical harmonic coefficients at each voxel replacing view direction information. 

Our proposed method is also based on the voxel grid representation and can be optimized directly by gradient back-propagation. What sets it apart is that our algorithm can dynamically enlarge certain voxel resolutions to fit local scene complexity during optimization, without requiring any pretrained geometric priors, unlike~\cite{fridovich2022plenoxels,yu2021plenoctrees}. Compared to the coarse-to-fine manner of DirectVoxGO~\cite{sun2022direct, sun2022improved}, it is more efficient in memory.

%% file: method.tex



In this work, we represent the 3D scene by using a dynamically partitioned grid structure. With such a representation, complex areas can be adaptively approximated by finer-scale voxels in iteration, while leaving well-fitted voxels unchanged. Moreover, with modalities of interest explicitly in its grid cells, we conduct reconstruction by jointly optimizing values stored on each vertex under the self-supervision of both color $\mathcal{I}$ and depth observations $\mathcal{D}$. Camera poses $\mathcal{T}$ can be acquired by SLAM or SfM. As illustrated in Fig.~\ref{fig:method}, we query arbitrary 3D radiance samples directly through interpolation based on current voxel partitioning, and dynamically update the grid representation according to the voxel split metrics in Sec.~\ref{subsec:optimization}.


\subsection{Volumetric Rendering in Voxel Grid}
A voxel grid representation approximates the scene geometry by a uniform 3D grid with different modalities of interest stored in each voxel node. For our RGB-D fusion task, the modalities correspond to SDF and radiance values. With such representation, it is efficient to query for arbitrary 3D positions $\mathbf{x}$ via interpolation: 
\begin{gather}
	s = interp(\mathbf{x}, \mathbf{G}^{(s)}(\mathbf{x})), \\
	\mathbf{r} = interp(\mathbf{x}, \mathbf{G}^{(r)}(\mathbf{x})),
\end{gather}
where $s$ and $\mathbf{r}$ are the SDF and radiance value respectively. $\mathbf{G} \in \mathbb{R}^{C \times N_{x} \times N_{y} \times N_{z}}$ is the voxel grid and $\mathbf{G}(\mathbf{x})$ returns the voxel in which $\mathbf{x}$ is allocated. $C$ is the dimension of modalities and $(N_{x}, N_{y}, N_{z})$ indicates the grid resolution. Typically, higher resolutions contribute to better representation ability. In our implementation, we utilize trilinear interpolation due to the trade-off between simplicity and stability.

Leveraging this property, it is thus easy to integrate volumetric rendering. Along each viewing ray, $N_k$ points are sampled and optimized for their corresponding SDF and radiance. We approximate the color along the viewing ray as a weighted sum of these samples:
\begin{equation}
    \mathbf{R} = \frac{1}{\sum^{N_k-1}_{i=0}  w_i} \sum^{N_k-1}_{i=0}  w_i \cdot \mathbf{r}_i,
    \label{eq:rendering}
\end{equation}
where $w$ is the weight gauging the visibility of current sample on the ray given by  $w_i = T_i\alpha_i$. $T_i$ is the \textit{accumulated transmittance} as in \cite{wang2022go}:
\begin{equation} \label{volume_rendering_T}
    T_i = \prod_{j=1}^{i-1}(1-\alpha_j),
\end{equation}

\begin{equation} \label{volume_rendering_alpha}
    \alpha_i = \max \bigg(\frac{\sigma_b(\phi(\mathbf{x}_{i})) - \sigma_b(\phi(\mathbf{x}_{i+1}))}{\sigma_b(\phi(\mathbf{x}_{i}))}, 0 \bigg),
\end{equation}
$\sigma_b(x) = (1 + e^{-bx})^{-1}$ controls the transition smoothness and $b$ is a learnable parameter.

The predicted color serves as an important auxiliary to depth regulations. Note that in contrast to NeRF models~\cite{azinovic2022neural,mildenhall2021nerf,sun2022direct} which need another round of optimization by incorporating view direction, we solely focus here on exploiting only view-invariant colors. We found that it is able to generate similar reconstructed geometries under sufficient grid resolutions. This modification contributes a further boost to our runtime and makes the entire optimization independent of any MLP components. 

\subsection{Hierarchical Representation}
\label{Sec: VoxelTrace}
A big limitation of the voxel grid structure is its cubic memory consumption. We address this by introducing a hierarchical structure to avoid excessively partitioning in regions that are able to be well-represented in their present resolutions. Additionally, the saved resource can be further leveraged to benefit those parts requiring finer cells in the following iteration. 

For the sake of clarity, we denote our hierarchically split grid as $\mathbf{H}$. It allows the 3D points bounded by different-sized voxels. For an arbitrary 3D query, we first locate its terminal node that has the finest split. Since now it is unable to trace it directly from the coordinate, therefore a recursive search is needed. Next, after getting the bounding voxel, modalities can be estimated via interpolation based on the distances to its eight vertices:
\begin{gather}
    \mathbf{V}_x = trace(\mathbf{x}, \mathbf{H}), \\
    \mathbf{m} = interp(\mathbf{x}, \mathbf{V}^{(m)}_x),
\end{gather}
where $\mathbf{V}$ is the terminal voxel of current query point, $\mathbf{m}$ corresponds to either the SDF or radiance value.

The grid is first initialized in the form of a uniform voxel grid $\mathbf{H}^{[0]} \in \mathbb{R}^{N_x \times N_y \times N_z}$ at a coarse resolution. During the optimization, it will be updated every $S_u$ steps by adaptively subdividing certain voxels into eight more cells, until all voxels are sufficiently split or reach the predefined deepest layer. 




\begin{figure}[!t]
    \centering
    \includegraphics[width=0.85\columnwidth]{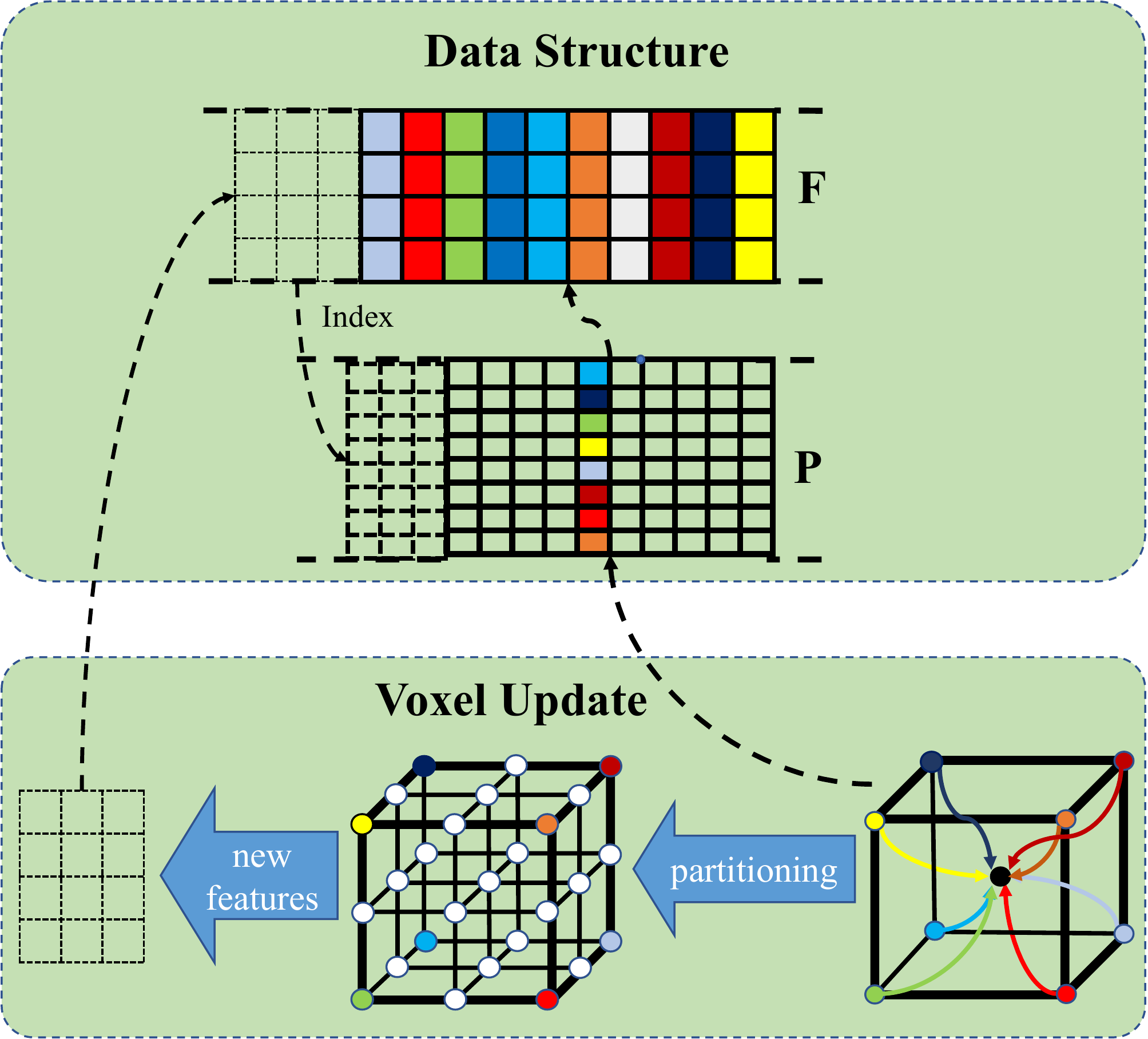}
    \caption{Each time a voxel being subdivided, new modalities and pointers will be added to $\mathbf{F}$ and $\mathbf{P}$ respectively. The original voxel will be deactivated and removed from $\mathbf{P}$.} 
	\label{fig:octree_detial}
\end{figure}

\noindent {\bf Data structure.}
Rather than naively storing the modalities in each voxel node, we gather them in a single matrix $\mathbf{F} \in \mathbb{R}^{N_{n}^{[l]} \times C}$, where $l \in [0, L_{max}]$ indicates current subdivision level, $N_{n}^{[l]}$ is the corresponding node number. Therefore, each node only needs to store a pointer directing to its modalities. We further group the pointers within each voxel, like Fig.~\ref{fig:octree_detial}, to form another matrix $\mathbf{P} \in \mathbb{R}^{N^{[l]} \times 8}$, collecting the information of entire grid. $N^{[l]}$ refers to the voxel number. Such vectorized implementation is more efficiency-friendly for training and traversing. 

In order to trace terminal voxels and their corresponding modalities, we additionally design a list $\mathbf{K}$ to record the structure variation. Initially, there is only one element in $\mathbf{K}$, \ie, $\mathbf{K}^{[0]} \in \mathbb{R}^{N^{[0]}}$ which stores the index of each voxel to $\mathbf{P}$. Whenever a new subdivision is performed, the data structures get updated accordingly by inserting new features into $\mathbf{F}$ and $\mathbf{P}$ and pruning redundancy in $\mathbf{P}$, \ie, the deactivated voxel as illustrated in~\cref{fig:octree_detial}. In the meantime, a new $\mathbf{K}^{[l]} \in \mathbb{R}^{N^{[l]}}$ is generated by performing a voxel indexing. More specifically, for every newly generated voxel, we calculate its midpoint $\mathbf{x}$ and assign it a unique node key with respect to $l$ as:
\begin{equation}
    key^{[l]} = (\overline{x}_2 \times 2^{l}N_x \times N_y + \overline{x}_1 \times 2^{l} N_x + \overline{x}_0),
    \label{eq: xyz2key}
\end{equation}
with $\mathbf{K}(key)$ storing the index to $\mathbf{P}$.

\noindent {\bf Voxel tracing.}
By virtue of this design, we can extract the corresponding terminal corner modalities of arbitrary node keys through a sequence of indexing operations as:
\begin{equation}
\begin{aligned}
    (s, \mathbf{r}) = \mathbf{F}(\mathbf{P}(\mathbf{K}^{[l]}(key^{[l]}))), \\
    \wrt, \mathbf{K}^{[l + 1]}(key^{[l + 1]}) == None.
\end{aligned}
\end{equation}
Since there usually are more than two levels, we thus trace the terminal voxels in a recursive manner.

\begin{figure*}[ht]
    \centering
	\includegraphics[width=0.9\textwidth]{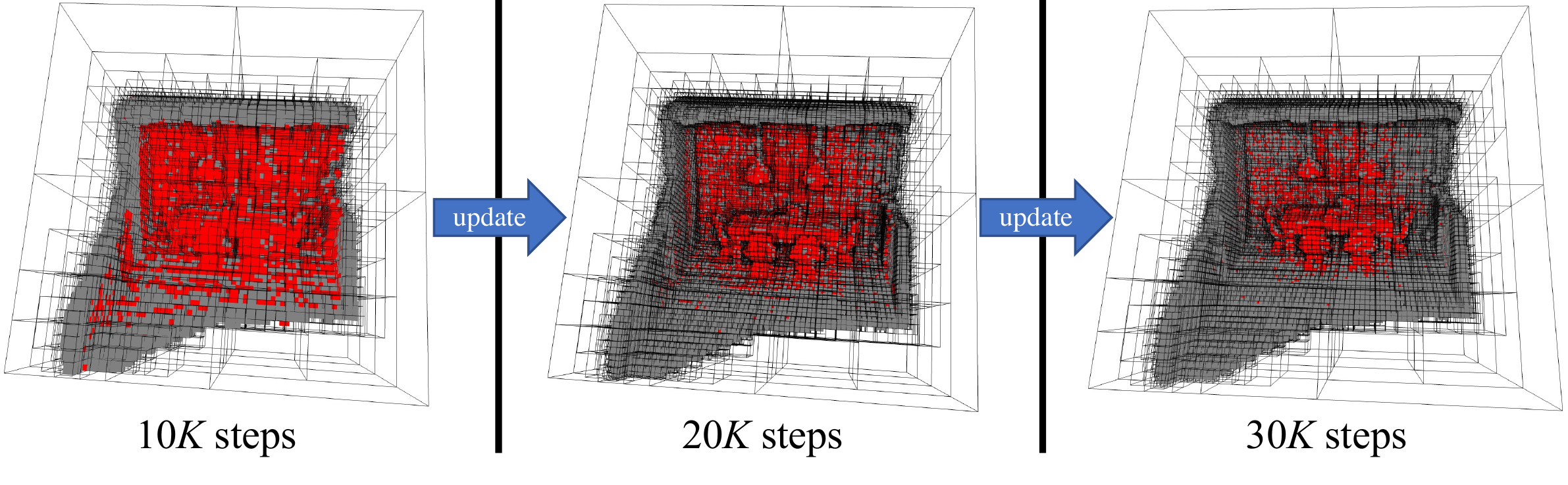}
	\caption{Our proposed updating strategy focuses on identifying and subdividing voxels in areas of high loss, which are typically indicative of complex structures. As the scene becomes better trained, our approach automatically narrows the updating area, prioritizing thin objects and detailed surfaces to avoid wasting computational resources on unnecessary areas. } 
	\label{fig:update}
\end{figure*}

\subsection{Optimization}
\label{subsec:optimization}
We first estimate the bounding box based on the camera frustums of input data. Then we randomly sample the pixels $\mathbf{I}$ with a certain batch size from the input RGB images and sample 3D points $\mathbf{X}$ along viewing ray between 
intersects with the bounding box. We directly use grayscale images instead of full RGB values, as it would weaken the impact from non-Lambertian materials and it is more computational-friendly.

With all 3D points sampled, we optimize our grid structure by minimizing the objective function with respect to node modalities $\mathbf{F}$ and camera poses $\mathcal{T}$. We define the truncation distance as $tr = 5cm$ and scale the scene to bound the truncation region in range $[-1, 1]$. 

\noindent {\bf Loss function.}
Out object function consists of three loss terms: image loss, free space loss, and SDF loss. The image loss measures the pixel-wise color difference between the rendered image and the ground-truth image:
\begin{equation}
    L_{rgb}(\mathbf{I}) = \frac{1}{|\mathbf{I}|} \sum_{i \in \mathbf{I}} (\mathbf{R}_i - \hat{\mathbf{R}}_i)^2,
    \label{loss:rgb}
\end{equation}
where $\hat{\mathbf{R}}$ represents the observed image. The free space loss term enforces the SDF value between the truncation region and the camera origin to be $tr$ as follows:
\begin{equation}
    L_{fs}(\mathbf{X}) = \frac{1}{|\mathbf{I}|} \sum_{i \in \mathbf{I}} \frac{1}{|\mathbf{X}^{fs}_i|} \sum_{\mathbf{x} \in \mathbf{X}^{fs}_i} (s_{\mathbf{x}} - tr)^2,
    \label{loss:fs}
\end{equation}
where $\mathbf{X}^{fs}$ denotes the set of samples belonging to the free space. For the samples lay in the truncation area, we use $L_{sdf}$ to encourage estimated SDF values to be equal to the ones computed from depth observations:
\begin{equation}
    L_{sdf}(\mathbf{X}) = \frac{1}{|\mathbf{I}|} \sum_{i \in \mathbf{I}} \frac{1}{|\mathbf{X}^{sdf}_i|} \sum_{\mathbf{x} \in \mathbf{X}^{sdf}_i} (s_{\mathbf{x}} - \hat{s}_{\mathbf{x}})^2.
    \label{loss:sdf}
\end{equation}
where $\mathbf{X}^{sdf}$ denote samples belonging to the truncated region. For the weight applied to each loss, please kindly refer to the supplementary material.

\noindent {\bf Dynamic partitioning.}
\label{Sec: update}
After iterating some steps, we review the representing capability of our voxel grid and perform a further subdivision for certain voxels if necessary. Our dynamic partitioning metrics are subject to two criteria: 

1) We limit the selection of voxels for subdivision to those lying between the truncation area. Specifically, we regard a voxel as a candidate for subdivision only if its eight corner SDF values satisfy the condition that the minimum value is smaller than $1$ and the maximum value is larger than $0$. This metric ensures that the subdivision process focuses on voxels that are more likely to contain relevant information and contribute to the overall quality increase.

2) Our partitioning is aimed at improving the optimization process by prioritizing voxels that have a higher probability to lower the holistic loss. We achieve this by initializing two matrices of the same size as the voxel number to store the image loss and truncation loss. During training, for each sampled 3D point, we update the corresponding matrix based on the voxel it lies in and the loss value. When subdividing, we select those voxels with either image loss or SDF loss greater than a threshold. It is reasonable to believe that a high image loss is more likely due to an insufficient resolution for representing the current texture, as is the case with the SDF loss. 

As illustrated in \cref{fig:update}, the proposed voxel refinement technique selectively subdivides voxels with higher loss values, labeled in color red, to achieve higher resolution representation. With iterating, the scheme gradually narrows the area and focuses on some really challenging parts. 

%% file: experiments.tex

We validated the performance of our method on both synthetic and real datasets. We first presented the quantitative evaluation results on the synthetic dataset to demonstrate the accuracy of our reconstruction. We then showed the advantage of our method in reconstructing fine details on both synthetic and real datasets. Finally, we provided experimental results that demonstrate the efficiency of our method, as well as an ablation study to analyze the contribution of each key component in our method. More implementation details and parameter settings can be found in the appendix. 

\subsection {Synthetic Dataset}
We conducted a comprehensive quantitative evaluation of our method using $10$ synthetic scenes provided by \cite{azinovic2022neural}. To simulate realistic conditions, noise and artifacts were added to depth observations and camera poses. 

We compare our approach with baseline methods, such as NeuralRGB-D \cite{azinovic2022neural} and GO-Surf \cite{wang2022go}, using four different evaluation metrics: the Chamfer $\ell_1$ distance, normal consistency, intersection-over-union (IoU) and the F\mbox{-}score. As demonstrated in \cref{tab:quantitative},  our method outperforms all the others in both Chamfer $\ell_1$ distance and IoU. While our our F-score of is slightly lower than that of NeuralRGBD, the overall performance of our method is still superior. We observed that GO-Surf achieves the best normal consistency, which can be attributed to its enforcement of normal smoothness during training.

\begin{figure*}[t!]
    \centering
	\includegraphics[width=\textwidth]{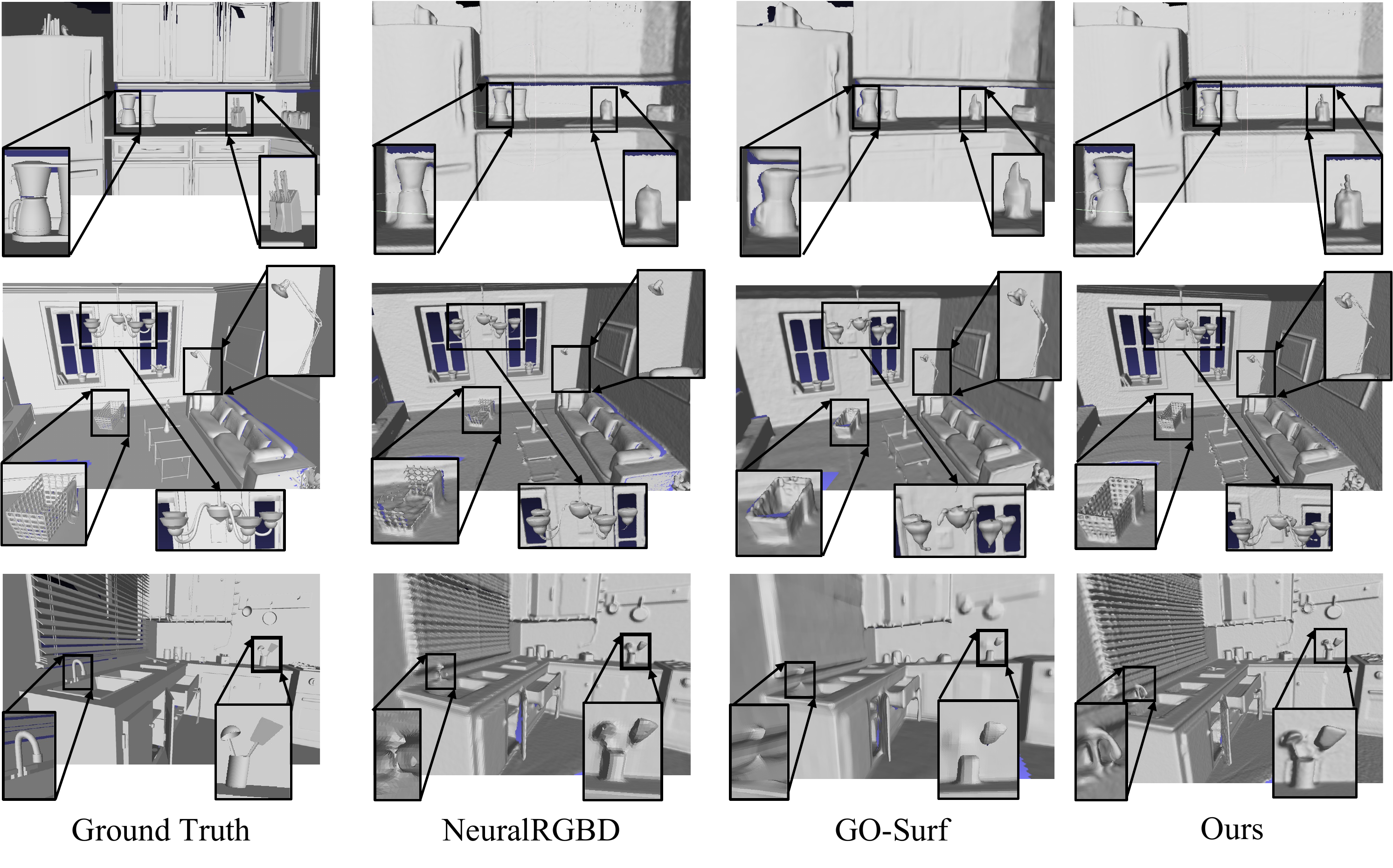}
	\caption{We compare our method with two SOTA techniques, namely NeuralRGBD \cite{azinovic2022neural} and GO-Surf \cite{wang2022go} on three scenes - Complete Kitchen, Green Room, and Morning Apartment - using 10 synthetic scenes provided by \cite{azinovic2022neural}. Our qualitative results demonstrate that our method outperforms both existing methods in terms of reconstruction completion and detail recovery.} 
	\label{fig:synthiic}
\end{figure*}

Furthermore, we provide visual comparisons of the reconstructed 3D scenes using our method, NeuralRGBD, and GO-Surf in \cref{fig:synthiic}. Notable that the proposed method shows superiority in terms of recovering fine details and complex structures, as shown in the closeup figures. While GO-Surf exhibits the most completed meshes, it fails to recover complicated structures due to its smoothness enforcement.

\begin{table}
	\resizebox{\linewidth}{!}{%
    \centering
    \begin{tabular}{lcccc}
        \toprule
        \textbf{Method}  & \textbf{C-$\ell_1$} $\downarrow$  & \textbf{IoU} $\uparrow$ & \textbf{NC} $\uparrow$  & \textbf{F-score} $\uparrow$ \\
        \midrule
        BundleFusion\cite{dai2017scannet}     & 0.062                             & 0.594                   & 0.892                   & 0.805	                   \\
        RoutedFusion\cite{weder2020routedfusion}     & 0.057                             & 0.615                   & 0.864                   & 0.838                       \\
        COLMAP\cite{schonberger2016structure,schonberger2017vote,schonberger2016pixelwise} + Poisson\cite{kazhdan2013screened} & 0.057                             & 0.619                   & 0.901                   & 0.839                       \\
        Conv. Occ. Nets\cite{peng2020convolutional}  & 0.077                             & 0.461                   & 0.849                   & 0.643                       \\
        SIREN\cite{sitzmann2020implicit}            & 0.060                             & 0.603                   & 0.893                   & 0.816                       \\
        NeRF\cite{mildenhall2021nerf} + Depth     & 0.065                             & 0.550                   & 0.768                   & 0.782                       \\
        NeuralRGBD\cite{azinovic2022neural}      & 0.044                             & 0.747                   & 0.918                   & \textbf{0.924}                       \\
        GO-Surf\cite{wang2022go}           & 0.041                             & 0.916          & \textbf{0.920}          & 0.907                       \\
        \midrule
        
        Ours             & \textbf{0.037}                   & \textbf{0.923}                  & 0.902                   & 0.920              \\
        \bottomrule
    \end{tabular}
    }
    \caption{To evaluate the reconstruction performance of our method, we conduct a quantitative analysis on 10 synthetic scenes provided by \cite{azinovic2022neural}. We directly cite the results reported in \cite{azinovic2022neural} and compare the performance of GO-Surf \cite{wang2022go} and our method under the same metrics. We compute the Chamfer $\ell_1$ distance, normal consistency, and the F\mbox{-}score~\cite{10.1145/3072959.3073599} between point clouds sampled with a density of $1$ point per $cm^{2}$, with a threshold of $5cm$ for the F-score. Moreover, we voxelize the mesh to compute the intersection-over-union (IoU) between predictions and ground truth. }
\vspace{-0.2cm}
\label{tab:quantitative}
\end{table}

\subsection {Real Dataset}
\begin{figure*}[t!]
    \centering
	\includegraphics[width=\textwidth]{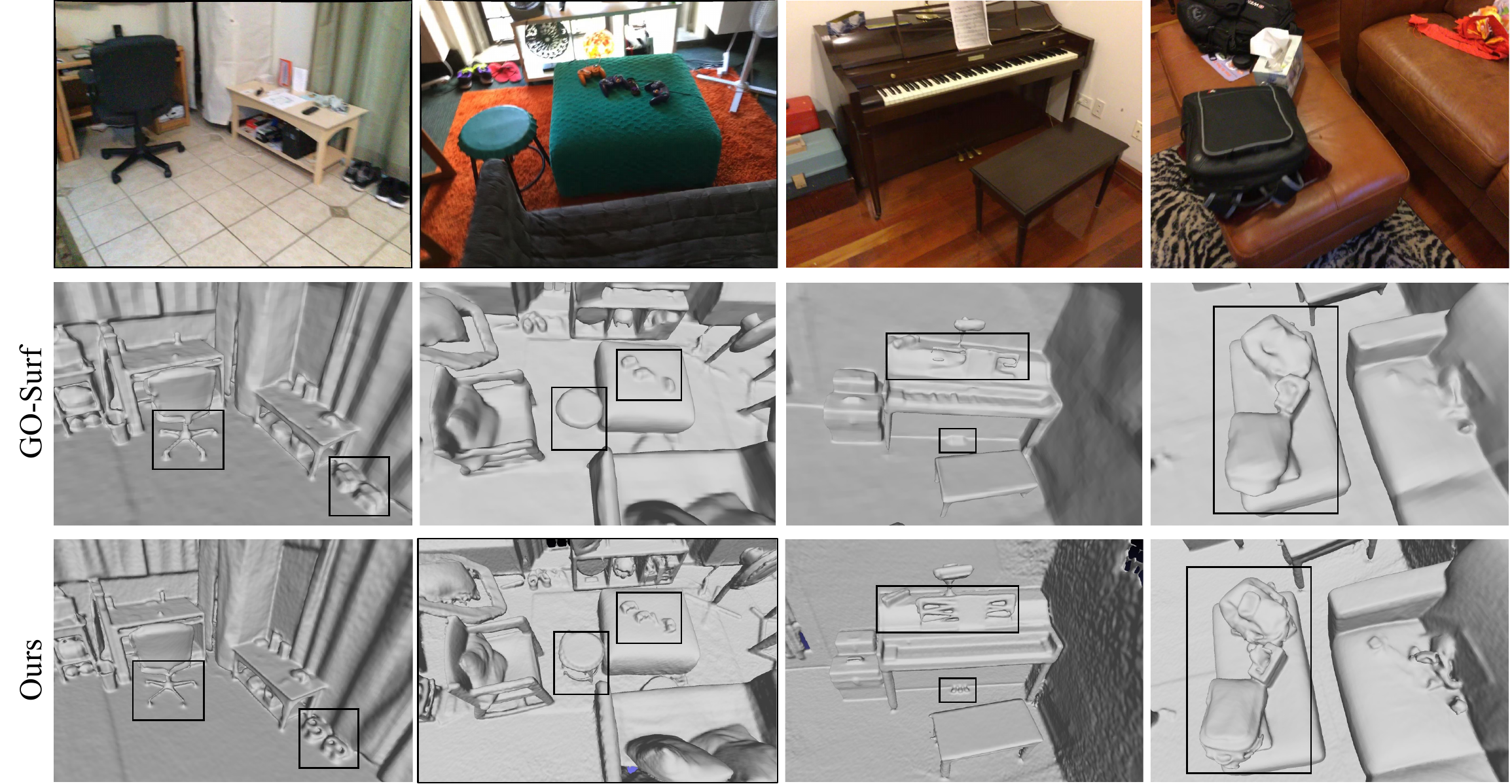}
	\caption{We conducted a comparative evaluation of our method and GO-Surf \cite{wang2022go} on scenes 1, 2, and 50 from the ScanNet dataset \cite{dai2017scannet}. The results demonstrate that our method is highly effective in recovering thin objects and complex surfaces, while GO-Surf often over-smooths the surfaces. } 
	\label{fig:scannet}
\end{figure*}

We also conducted experiments on the real-world RGB-D dataset ScanNet \cite{dai2017scannet}, which presents difficulties due to noisy depth, missing observations, and image blur. \cref{fig:scannet} exhibits the reconstruction results from the proposed method and GO-Surf. Similar to the synthetic scenes, our method demonstrates good performance in recovering complex structures and details of objects, such as the shape of shoes, wrinkles on bags, and laces around chairs. In contrast, GO-Surf emphasizes smoother surfaces at the cost of losing finer details. More comparison results can be found in the appendix. Our results suggest that the proposed method is well-suitable for real-world applications, where the accurate reconstruction of complex objects with fine details is crucial.

\subsection {Ablaition Study}
\begin{figure*}[t!] 
    \centering
    \includegraphics[width=0.95\textwidth]{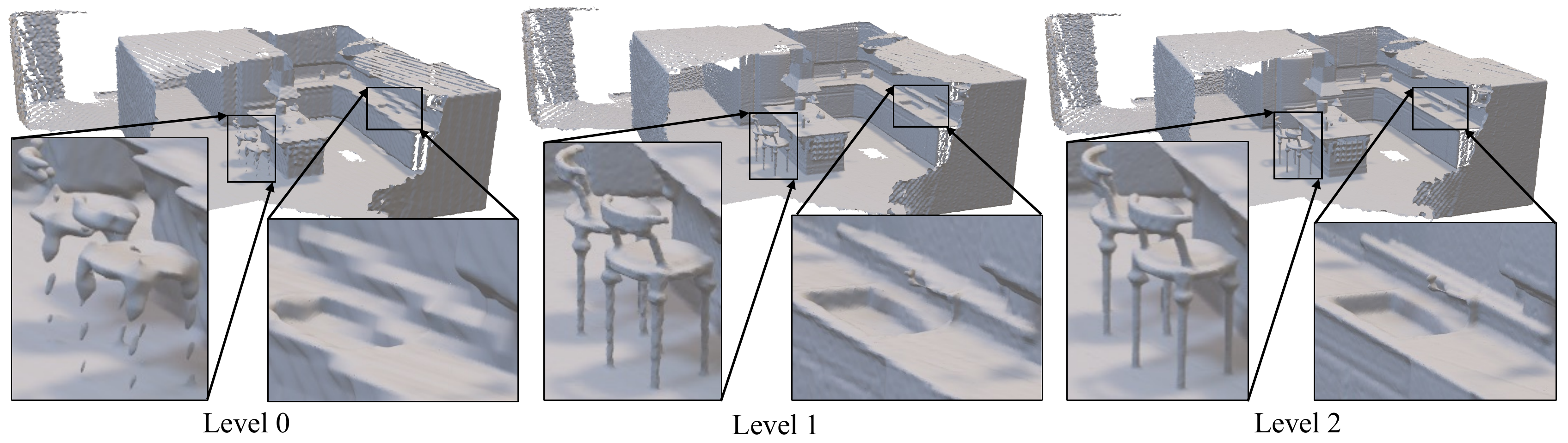}
    \caption{We present the reconstruction results obtained with varying subdivision level of our voxel grid. We demonstrate that increasing the subdivision level of the structure results in superior reconstruction quality.}
    \label{fig:difdepth}
    \vspace{-5pt}
\end{figure*}

\begin{table}
	\resizebox{\linewidth}{!}{%
    \centering
    \begin{tabular}{lcccc}
        \toprule
        \textbf{Method}  & \textbf{C-$\ell_1$} $\downarrow$  & \textbf{IoU} $\uparrow$ & \textbf{NC} $\uparrow$  & \textbf{F-score} $\uparrow$ \\
        \midrule
    Ours(surf+loss)     & 0.03654                             & 0.923                       & 0.9017                   & 0.9195                  \\
        Ours(surf)          & 0.03836                             & 0.9229                   & 0.9013                   & 0.9196                       \\
        \bottomrule
    \end{tabular}
    }
    \caption{Experiment results updating on the surface with large loss value and updating on the whole surface.}
\vspace{-0.2cm}
\label{tab:ablation_update}
\end{table}

\begin{table}
	\resizebox{\linewidth}{!}{%
    \centering
    \begin{tabular}{lccccc}
        \toprule
        \textbf{Method}  & \textbf{C-$\ell_1$} $\downarrow$  & \textbf{IoU} $\uparrow$ & \textbf{NC} $\uparrow$  & \textbf{F-score}  $\uparrow$  & \textbf{PSNR} \\
        \midrule
    Ours($64^3$, grayscale)     & 0.03921                             & 0.9170                       & 0.9136                   & 0.9103       &  27.59           \\
        Ours($64^3$, rgb)          & 0.03920                             & 0.9119                   & 0.9143                   & 0.9103       &  26.94              \\
        Ours($64^3$, grayscale+SH)          & 0.03905                            & 0.9255                  & 0.9125                  & 0.9102     &30.55                 \\
        Ours($64^3$, rgb+SH)          & 0.03929                           & 0.9212                 & 0.9143                  & 0.9098    &29.61                 \\
        \bottomrule
    \end{tabular}
    }
    \caption{Results rendering with different color strategies.}
\vspace{-0.2cm}
\label{tab:ablation_color}
\end{table}

\noindent {\bf Voxel subdivision level.}
We first investigated the impact of different subdivision levels of voxel grid on the quality of reconstructed 3D scenes. We initialized the voxel grid with a resolution of $128^3$ and train it for $30K$ steps. To examine the effect of the subdivision level, we performed three experiments. In the first experiment, we did not subdivide the voxel during the entire training process, so the subdivision level $L_{max} = 1$. In the second experiment, we subdivided the voxel once at 10K steps, resulting in $L_{max} = 2$. Finally, in the third experiment, we subdivided the voxel at $10K$ and $20K$ steps, leading to $L_{max} = 3$.

The reconstruction results shown in \cref{fig:difdepth} demonstrate that a higher level of subdivision contributes to better reconstruction quality. Specifically, the model trained with a non-subdivided structure exhibits the lowest quality, with missing structures and coarse details. 


\noindent {\bf Voxel subdivision criterion.}
We conducted an extensive investigation into the impact of our voxel subdivision strategies on the proposed method. Specifically, we compared the performance of subdividing the voxels naively on the entire surface versus selectively in areas with larger loss values, as described in~\cref{Sec: update}. Our experimental results, summarized in~\cref{tab:ablation_update}, demonstrate that the reconstruction results are comparable between the two strategies. This indicates that our strategy, which focuses on high-loss areas, is effective in reconstructing surfaces.

\noindent {\bf Image color.}
Here we investigated the impact of color formats on our method. Specifically, we compared the performance of four models which are initialized with $64^3$ resolution. The first two are trained with view-independent grayscale and RGB values directly stored in voxel nodes. Another two are computed based on spherical harmonic coefficients (SH), as \cite{yu2021plenoctrees}, in order to preserve view-dependent information. We keep all other parameters constant to isolate the effect of different color strategies on reconstruction accuracy. As illustrated in~\cref{tab:ablation_color}, the grayscale format shows an overall balance among the five metrics. Moreover, while the view-dependent information helps to yield higher peak signal-to-noise ratio (PSNR) scores, it does not significantly affect the quality of 3D reconstruction.

\noindent {\bf Initial resolution.}
We further performed two tests to investigate the effect of initialization resolution on our structure. They are initialized respectively using $256^3$ and $64^3$ and running for $60K$ steps. The structures were then updated separately by twice and four times, to reach the same resolution level. The results, presented in \cref{tab:quantitative} and \cref{tab:ablation_color}, show that our approach, denoted as $Ours$, outperforms the variant initialized with lower resolution, $Ours(64^3, grayscale)$. Initializing with lower resolution results in a deeper structure, leading to a longer search time for terminal nodes. Conversely, the lower resolution initialization would provide a more memory-efficient solution by leaving most free-space voxels undivided, which could consequently lead to fewer total voxels.
\begin{table}
	\resizebox{\linewidth}{!}{%
    \centering
    \begin{tabular}{lccccc}
        \toprule
        \textbf{Method}  & \textbf{level 0} & \textbf{level 1} & \textbf{level 2}  & \textbf{level 3}  & \textbf{level 4} \\
        \midrule
        Dynamic voxel grid     & 3,747MB                             & 4,249MB                   & 5,967MB                  & 12,771MB                  & 32,023MB	                   \\
        Voxel grid      & 3,245MB                             & 4,263MB                   & 10,983MB                 & OOM                    & OOM      \\
        \bottomrule
    \end{tabular}
    }
    \caption{GPU memory consumption between full-size and dynamize voxel grid. Our structure is more GPU-efficient. OOM stands for out-of-memory}
\vspace{-0.2cm}
\label{tab:MemoEff}
\end{table}

\subsection {Memory Efficiency} 
Our proposed method provides a distinct advantage over voxel-based methods \cite{sun2022direct, sun2022improved} so that we can refine local areas without increasing the resolution of the entire space. This approach allows us to extract more information and improve reconstruction within the same memory limit by focusing on areas with the most potential for refinement. To demonstrate the efficiency of our dynamic voxel grid structure, we reconstructed \textit{complete kitchen} in the synthetic dataset, starting at a resolution of $128^3$. We performed the dynamic partitioning every 10$K$ steps until the CUDA memory ran out. As shown in \cref{tab:MemoEff}, directly increasing the entire grid resolution only goes to level $2$ before running out of memory. In contrast, our method can keep updating until reaching level $4$, which has more chances to represent finer structures. For runtime, details can be found in the appendix.

%% file: conclusions.tex
In this work, we have presented a new RGB-D surface reconstruction method by exploring dynamic voxel grid optimization. It takes both color and depth observations as guidance. By directly storing and regressing modalities on grid voxels, the entire reconstruction converges fast and requires no MLP structures. Meanwhile,
our selective subdivision strategy can dynamically represent complex objects during optimizaiton without any pretrained geometric priors, refraining from naively holistic resolution expansion. The comprehensive evaluation demonstrates the effectiveness of our approach in generating high-quality 3D reconstructions with fine details, while maintaining computational efficiency. 

%% file: appendix.tex
\section{Implementation Details}


\subsection{Voxel Grid Initialization}


We initialize our voxel grid $\mathbf{H}^{[0]} \in \mathbb{R}^{N_x \times N_y \times N_z}$ proportional to the size of the bounding box, where the total voxel number $N_x \times N_y \times N_z$ is set to $256^3$. For initial modalities values on each node, we set the boundary SDF values of the grid to $0$, while assigning a value of $1$ to the internal nodes. Radiance values are all set to $0$ in the range of $[0, 1]$ when the optimization begins.

\begin{figure}[t!]
    \centering
	\includegraphics[width=0.5\textwidth]{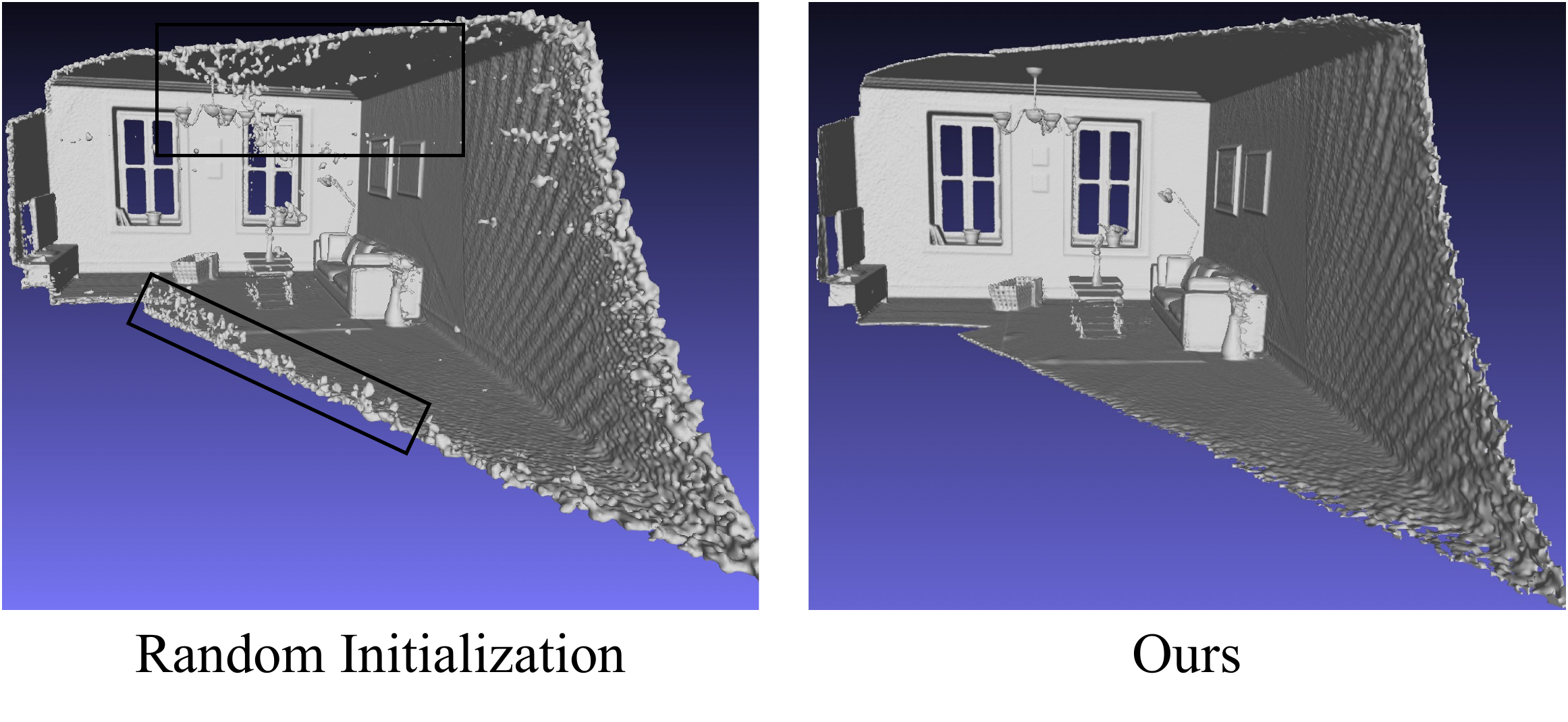}
	\caption{The comparison of different initialization solutions.} 
	\label{fig:inital}
\end{figure}

We show a comparison of reconstructed results by respectively initializing the SDFs with random values from $-1$ to $1$ and using our strategy in \cref{fig:inital}. This easy process can boost the convergence and significantly constrict outliers introduced around reconstructed surfaces. 


\subsection{Objective Function}
Our objective function integrates three distinct loss terms: image loss, free space loss, and SDF loss as follows:
\begin{equation}
    L(\mathbf{I}, \mathbf{X}) = \lambda_1 L_{rgb}(\mathbf{I}) + \lambda_2 L_{fs}(\mathbf{X}) + \lambda_3 L_{sdf}(\mathbf{X}),
\end{equation}
where all of these terms have been introduced in Sec. 3.3 of the main paper. In our experiments, we set the weights respectively to $\lambda_1 = 2$, $\lambda_2 = 1$, and $\lambda_3 = 0.2$.

\subsection{Optimization Setting}
Without being specifically mentioned, we train our model for a total of $60K$ iterations, with voxel partitioning performed twice at $20K$ and $40K$ iterations respectively. For voxels lying in the truncation area, we select them as candidates if either the estimated SDF loss or image loss is greater than $0.0001$.

The learning rate is initialized to $0.1$, with an exponential learning rate decay of $0.1$ applied at $20K$ iterations. We do employ pose refinement as \cite{azinovic2022neural} but no deformation step is applied, as we note very limited improvement in our cases.

In each iteration, we randomly select $2,048$ rays and uniformly sample $128$ points on each ray within the bounding box by adding a small random perturbation. We then sample another $128$ points, based on the estimated visibility weights $\omega$ in Eq. 3 of the main paper, through one step \cite{wang2021neus}. 

\section{More Results}
\begin{figure*}[t!]
    \centering
	\includegraphics[width=0.85\textwidth]      {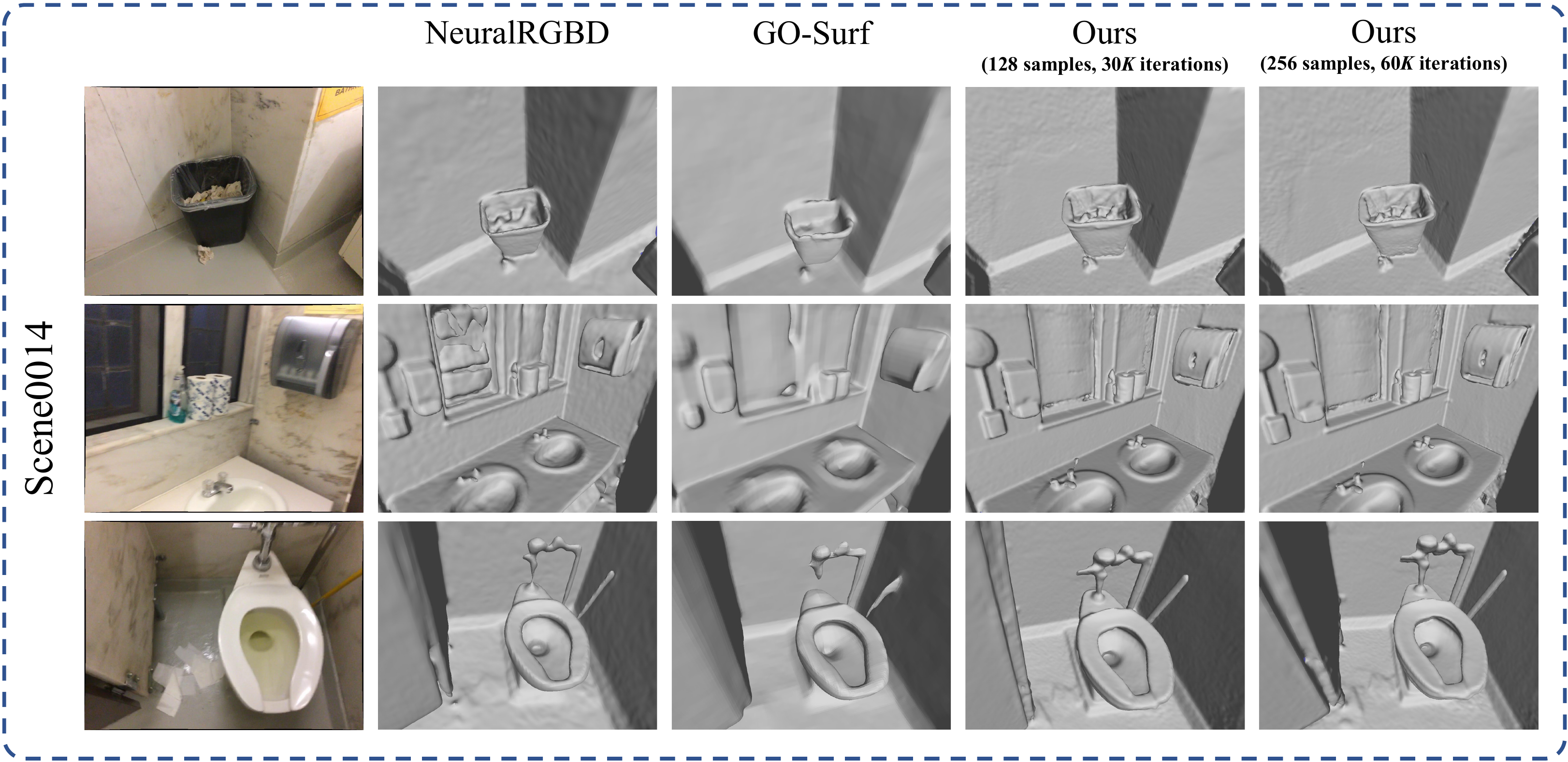}
        \includegraphics[width=0.85\textwidth]{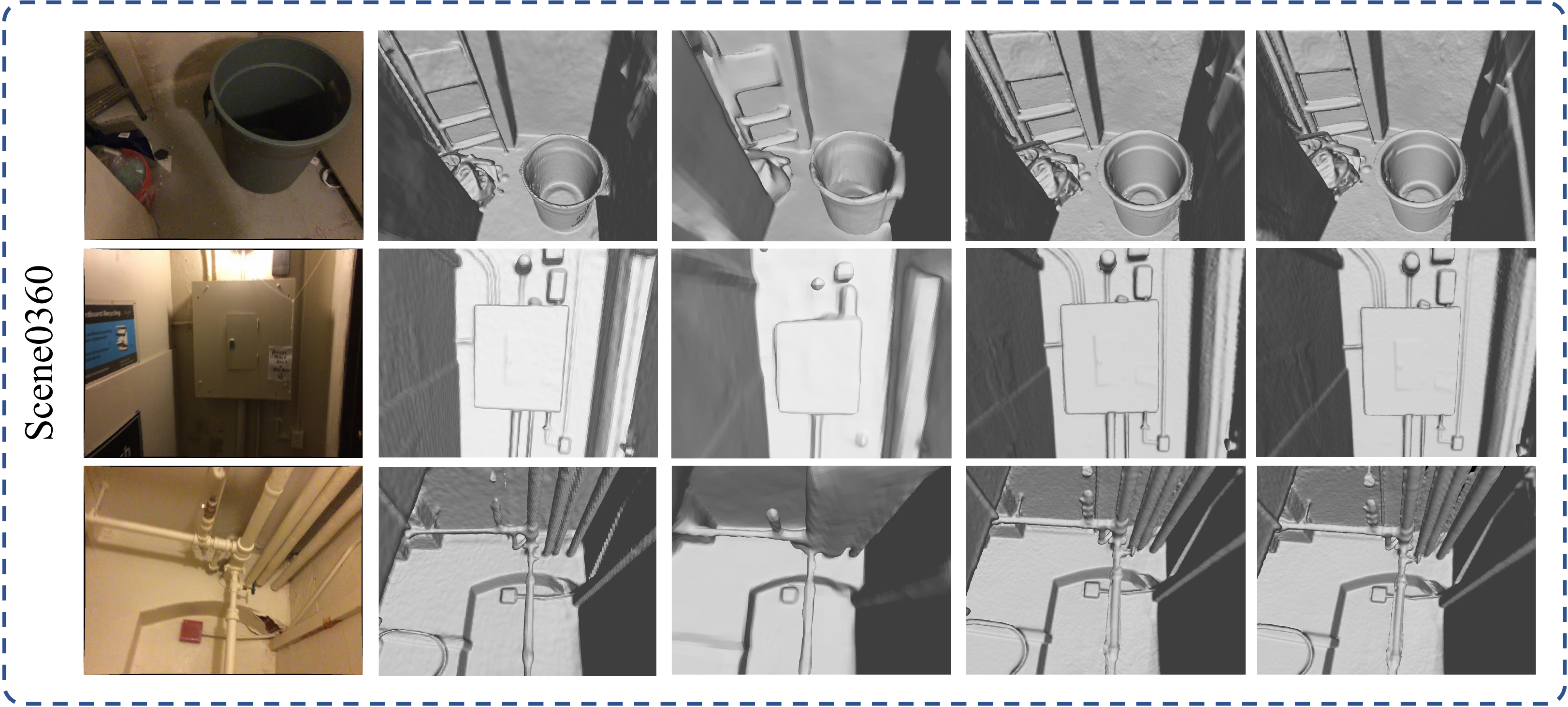}
        \includegraphics[width=0.85\textwidth]{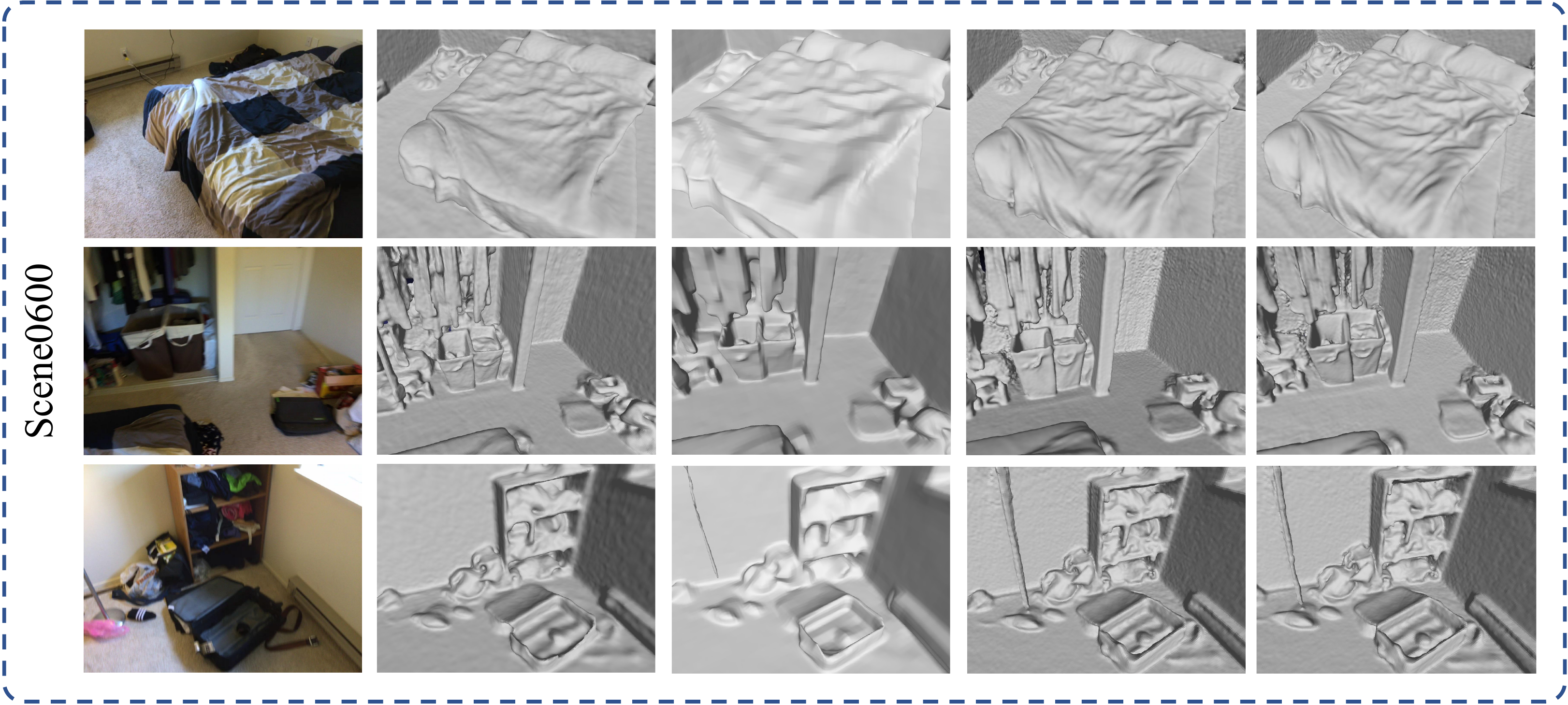}
	\caption{To evaluate the quality of our method, we performed a comparative visualization with neuralRGBD \cite{azinovic2022neural} and GO-Surf \cite{wang2022go} on three different scenes (0014, 0360, and 0600) from the ScanNet dataset \cite{dai2017scannet}.  Our method shows superiority in recovering high-fidelity geometric details.} 
	\label{fig:scannet2}
\end{figure*}

\begin{figure*}[t!]
    \centering
	\includegraphics[width=0.9\textwidth]{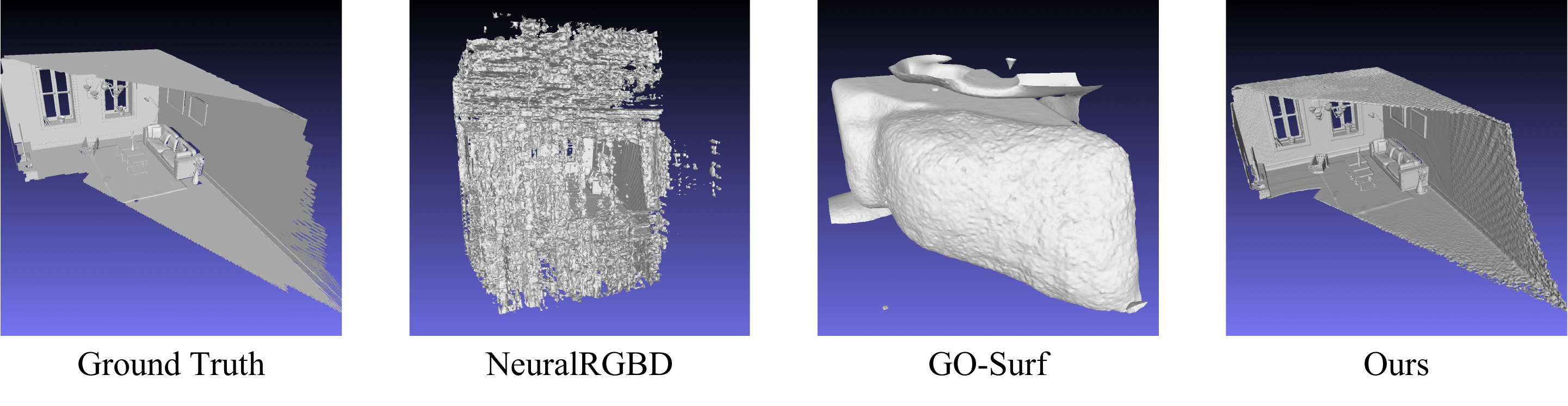}
	\caption{In terms of raw mesh results, it is able to observe distinct differences among NeuralRGBD \cite{azinovic2022neural}, GO-Surf \cite{wang2022go} and ours. The raw result generated by NeuralRGBD appears noisy without culling. GO-Surf exhibits a sphere-like shape due to its sphere initialization. Our raw result can be much closer to the ground truth without any post-processing.} 
	\label{fig:raw_mesh}
\end{figure*}

\label{Sec:ExpDetail}


To evaluate the performance of our approach, we first uniformly query the SDF values located inside the bounding box at the same resolution as $\mathbf{H}^{[0]} \in \mathbb{R}^{N_x \times N_y \times N_z}$ from our final dynamic voxel grid. The query step follows the voxel tracing pipeline in Sec. 3.2 of the main paper. Then, a triangle mesh can be recovered via marching cubes \cite{lorensen1987marching} based on the queried SDF values. Like~\cite{azinovic2022neural, wang2022go}, we \textit{cull} regions that are not observed by any cameras before evaluation. However, this process actually contributes to a limited increase for our models as discussed later. Our algorithm is implemented in PyTorch and runs on a single NVIDIA RTX A6000. For comparison purposes, we run NeuralRGBD~\cite{azinovic2022neural} and GO-Surf~\cite{wang2022go} in their default settings.

\subsection{Qualitative Results}

We provide more qualitative comparison results with NeuralRGBD \cite{azinovic2022neural} and GO-Surf \cite{wang2022go} on the ScanNet dataset \cite{dai2017scannet} as shown in \cref{fig:scannet2}. Notable that our approach demonstrates superior performance in recovering fine-grained structural details compared to the other two methods.

\subsection{Raw Mesh Comparison}

To illustrate the post-process reliance, we present the raw reconstruction results of NeuralRGBD \cite{azinovic2022neural}, GO-Surf \cite{wang2022go}, and our proposed method on the $Green Room$ scene \cite{azinovic2022neural} without \textit{culling} in \cref{fig:raw_mesh}. Our approach is able to yield a raw reconstruction result that is remarkably close to the ground truth without any post-processing. In contrast, the raw mesh of NeuralRGBD appears noisy without the post-processing to cull unobserved regions. GO-Surf presumes a sphere-like shape and reconstructs a closed surface, which may not accurately reflect the true geometry of the scene.

\subsection{Runtime}
\begin{table}
	\resizebox{\linewidth}{!}{%
    \centering
    \begin{tabular}{lcccc}
        \toprule
        \textbf{Method}  & \textbf{scene0014} & \textbf{scene0360} & \textbf{scene0600} & \textbf{average} \\
        \midrule
        NeuralRGBD\cite{azinovic2022neural}      & 20h14m19s                             & 20h10m21s                   & 20h6m51s                   & 20h10m30s                      \\
        GO-Surf\cite{wang2022go}           & 19m25s                             & 17m3s         & 20m45s          & 19m4s                     \\
        \midrule
        
        Ours             & 47m43s                   & 49m52s                  & 45m55s                   & 47m50s              \\
        Ours$^\ast$             & 22m33s                   & 21m28s                 & 21m41s                  & 21m54s              \\
        \bottomrule
    \end{tabular}
    }
    \caption{Comparison of running time between NeuralRGBD\cite{azinovic2022neural}, GO-Surf\cite{wang2022go}, and our method. Ours$^\ast$ represents the training cost after reducing the sampling number to $128$ and running iterations to $30K$, and subdividing at $10K$ and $20K$ iterations.}
\vspace{-0.2cm}
\label{tab:run_time}
\end{table}
As demonstrated in \cref{tab:run_time}, our proposed method offers a significant improvement in terms of training time compared to NeuralRGBD. In contrast to the hours of training time required by NeuralRGBD, our method needs only about $50$ minutes, making it much more affordable and scalable for real-world applications. Additionally, by reducing the total sampling number to $128$ and running iterations to $30K$, and subdividing at $10K$ and $20K$ iterations, our method can further reduce the training time to about $20$ minutes while maintaining high reconstruction quality as shown in~\cref{fig:scannet2}, which is comparable to the training time required by GO-Surf.
\section{Limitations}
The proposed approach also faces several limitations. First, while the method can recover more detailed 3D geometries, it is unable to fill holes that are not captured by images. Furthermore, the method may result in a pitted appearance on some reconstructed surfaces due to textureless and noisy depth observations. 


%% file: main.bbl
\begin{thebibliography}{10}\itemsep=-1pt

\bibitem{azinovic2022neural}
Dejan Azinovi{\'c}, Ricardo Martin-Brualla, Dan~B Goldman, Matthias
  Nie{\ss}ner, and Justus Thies.
\newblock Neural rgb-d surface reconstruction.
\newblock In {\em Proceedings of the IEEE/CVF Conference on Computer Vision and
  Pattern Recognition}, pages 6290--6301, 2022.

\bibitem{chen2022tensorf}
Anpei Chen, Zexiang Xu, Andreas Geiger, Jingyi Yu, and Hao Su.
\newblock Tensorf: Tensorial radiance fields.
\newblock In {\em ECCV}, pages 333--350, 2022.

\bibitem{curless1996volumetric}
Brian Curless and Marc Levoy.
\newblock A volumetric method for building complex models from range images.
\newblock In {\em Proceedings of the 23rd Annual Conference on Computer
  Graphics and Interactive Techniques}, pages 303--312, 1996.

\bibitem{dai2017scannet}
Angela Dai, Angel~X. Chang, Manolis Savva, Maciej Halber, Thomas Funkhouser,
  and Matthias Nie{\ss}ner.
\newblock Scannet: Richly-annotated 3d reconstructions of indoor scenes.
\newblock In {\em CVPR}, 2017.

\bibitem{dai2017bundlefusion}
Angela Dai, Matthias Nie{\ss}ner, Michael Zollh{\"o}fer, Shahram Izadi, and
  Christian Theobalt.
\newblock Bundlefusion: Real-time globally consistent 3d reconstruction using
  on-the-fly surface reintegration.
\newblock {\em ACM Transactions on Graphics (ToG)}, 36(4):1, 2017.

\bibitem{fridovich2022plenoxels}
Sara Fridovich-Keil, Alex Yu, Matthew Tancik, Qinhong Chen, Benjamin Recht, and
  Angjoo Kanazawa.
\newblock Plenoxels: Radiance fields without neural networks.
\newblock In {\em Proceedings of the IEEE/CVF Conference on Computer Vision and
  Pattern Recognition}, pages 5501--5510, 2022.

\bibitem{fu2020joint}
Yanping Fu, Qingan Yan, Jie Liao, and Chunxia Xiao.
\newblock Joint texture and geometry optimization for rgb-d reconstruction.
\newblock In {\em Proceedings of the IEEE/CVF Conference on Computer Vision and
  Pattern Recognition}, pages 5950--5959, 2020.

\bibitem{ji2022georefine}
Pan Ji, Qingan Yan, Yuxin Ma, and Yi Xu.
\newblock Georefine: Self-supervised online depth refinement for accurate dense
  mapping.
\newblock In {\em Computer Vision--ECCV 2022: 17th European Conference, Tel
  Aviv, Israel, October 23--27, 2022, Proceedings, Part I}, pages 360--377.
  Springer, 2022.

\bibitem{kazhdan2013screened}
Michael Kazhdan and Hugues Hoppe.
\newblock Screened poisson surface reconstruction.
\newblock {\em ACM Transactions on Graphics (ToG)}, 32(3):1--13, 2013.

\bibitem{10.1145/3072959.3073599}
Arno Knapitsch, Jaesik Park, Qian-Yi Zhou, and Vladlen Koltun.
\newblock Tanks and temples: Benchmarking large-scale scene reconstruction.
\newblock {\em ACM Trans. Graph.}, 36(4), July 2017.

\bibitem{laurentini1994visual}
Aldo Laurentini.
\newblock The visual hull concept for silhouette-based image understanding.
\newblock {\em IEEE Transactions on pattern analysis and machine intelligence},
  16(2):150--162, 1994.

\bibitem{li2022bnv}
Kejie Li, Yansong Tang, Victor~Adrian Prisacariu, and Philip~HS Torr.
\newblock Bnv-fusion: Dense 3d reconstruction using bi-level neural volume
  fusion.
\newblock In {\em Proceedings of the IEEE/CVF Conference on Computer Vision and
  Pattern Recognition (CVPR)}, 2022.

\bibitem{liu2020neural}
Lingjie Liu, Jiatao Gu, Kyaw Zaw~Lin, Tat-Seng Chua, and Christian Theobalt.
\newblock Neural sparse voxel fields.
\newblock {\em Advances in Neural Information Processing Systems},
  33:15651--15663, 2020.

\bibitem{lorensen1987marching}
William~E Lorensen and Harvey~E Cline.
\newblock Marching cubes: A high resolution 3d surface construction algorithm.
\newblock {\em ACM siggraph computer graphics}, 21(4):163--169, 1987.

\bibitem{maier2017intrinsic3d}
Robert Maier, Kihwan Kim, Daniel Cremers, Jan Kautz, and Matthias Nie{\ss}ner.
\newblock Intrinsic3d: High-quality 3d reconstruction by joint appearance and
  geometry optimization with spatially-varying lighting.
\newblock In {\em Proceedings of the IEEE international conference on computer
  vision}, pages 3114--3122, 2017.

\bibitem{mildenhall2021nerf}
Ben Mildenhall, Pratul~P Srinivasan, Matthew Tancik, Jonathan~T Barron, Ravi
  Ramamoorthi, and Ren Ng.
\newblock Nerf: Representing scenes as neural radiance fields for view
  synthesis.
\newblock {\em Communications of the ACM}, 65(1):99--106, 2021.

\bibitem{muller2022instant}
Thomas M{\"u}ller, Alex Evans, Christoph Schied, and Alexander Keller.
\newblock Instant neural graphics primitives with a multiresolution hash
  encoding.
\newblock {\em ACM Transactions on Graphics}, 41(4):1--15, 2022.

\bibitem{mur2015orb}
Raul Mur-Artal, Jose Maria~Martinez Montiel, and Juan~D Tardos.
\newblock Orb-slam: a versatile and accurate monocular slam system.
\newblock {\em IEEE transactions on robotics}, 31(5):1147--1163, 2015.

\bibitem{neff2021donerf}
Thomas Neff, Pascal Stadlbauer, Mathias Parger, Andreas Kurz, Joerg~H Mueller,
  Chakravarty R~Alla Chaitanya, Anton Kaplanyan, and Markus Steinberger.
\newblock Donerf: Towards real-time rendering of compact neural radiance fields
  using depth oracle networks.
\newblock In {\em Computer Graphics Forum}, volume~40, pages 45--59, 2021.

\bibitem{newcombe2015dynamicfusion}
Richard~A Newcombe, Dieter Fox, and Steven~M Seitz.
\newblock Dynamicfusion: Reconstruction and tracking of non-rigid scenes in
  real-time.
\newblock In {\em Proceedings of the IEEE conference on computer vision and
  pattern recognition}, pages 343--352, 2015.

\bibitem{newcombe2011kinectfusion}
Richard~A Newcombe, Shahram Izadi, Otmar Hilliges, David Molyneaux, David Kim,
  Andrew~J Davison, Pushmeet Kohi, Jamie Shotton, Steve Hodges, and Andrew
  Fitzgibbon.
\newblock Kinectfusion: Real-time dense surface mapping and tracking.
\newblock In {\em 2011 10th IEEE international symposium on mixed and augmented
  reality}, pages 127--136, 2011.

\bibitem{niemeyer2020differentiable}
Michael Niemeyer, Lars Mescheder, Michael Oechsle, and Andreas Geiger.
\newblock Differentiable volumetric rendering: Learning implicit 3d
  representations without 3d supervision.
\newblock In {\em Proceedings of the IEEE/CVF Conference on Computer Vision and
  Pattern Recognition}, pages 3504--3515, 2020.

\bibitem{niessner2013real}
Matthias Nie{\ss}ner, Michael Zollh{\"o}fer, Shahram Izadi, and Marc
  Stamminger.
\newblock Real-time 3d reconstruction at scale using voxel hashing.
\newblock {\em ACM Transactions on Graphics}, 32(6):1--11, 2013.

\bibitem{peng2020convolutional}
Songyou Peng, Michael Niemeyer, Lars Mescheder, Marc Pollefeys, and Andreas
  Geiger.
\newblock Convolutional occupancy networks.
\newblock In {\em Computer Vision--ECCV 2020: 16th European Conference,
  Glasgow, UK, August 23--28, 2020, Proceedings, Part III 16}, pages 523--540.
  Springer, 2020.

\bibitem{qin2018vins}
Tong Qin, Peiliang Li, and Shaojie Shen.
\newblock Vins-mono: A robust and versatile monocular visual-inertial state
  estimator.
\newblock {\em IEEE Transactions on Robotics}, 34(4):1004--1020, 2018.

\bibitem{reiser2021kilonerf}
Christian Reiser, Songyou Peng, Yiyi Liao, and Andreas Geiger.
\newblock Kilonerf: Speeding up neural radiance fields with thousands of tiny
  mlps.
\newblock In {\em Proceedings of the IEEE/CVF International Conference on
  Computer Vision}, pages 14335--14345, 2021.

\bibitem{schonberger2016structure}
Johannes~L Schonberger and Jan-Michael Frahm.
\newblock Structure-from-motion revisited.
\newblock In {\em Proceedings of the IEEE conference on computer vision and
  pattern recognition}, pages 4104--4113, 2016.

\bibitem{schonberger2017vote}
Johannes~L Sch{\"o}nberger, True Price, Torsten Sattler, Jan-Michael Frahm, and
  Marc Pollefeys.
\newblock A vote-and-verify strategy for fast spatial verification in image
  retrieval.
\newblock In {\em Computer Vision--ACCV 2016: 13th Asian Conference on Computer
  Vision, Taipei, Taiwan, November 20-24, 2016, Revised Selected Papers, Part I
  13}, pages 321--337. Springer, 2017.

\bibitem{schonberger2016pixelwise}
Johannes~L Sch{\"o}nberger, Enliang Zheng, Jan-Michael Frahm, and Marc
  Pollefeys.
\newblock Pixelwise view selection for unstructured multi-view stereo.
\newblock In {\em Computer Vision--ECCV 2016: 14th European Conference,
  Amsterdam, The Netherlands, October 11-14, 2016, Proceedings, Part III 14},
  pages 501--518. Springer, 2016.

\bibitem{sitzmann2020implicit}
Vincent Sitzmann, Julien Martel, Alexander Bergman, David Lindell, and Gordon
  Wetzstein.
\newblock Implicit neural representations with periodic activation functions.
\newblock {\em Advances in Neural Information Processing Systems},
  33:7462--7473, 2020.

\bibitem{snavely2006photo}
Noah Snavely, Steven~M Seitz, and Richard Szeliski.
\newblock Photo tourism: exploring photo collections in 3d.
\newblock In {\em SIGGRAPH}, pages 835--846. 2006.

\bibitem{sun2022direct}
Cheng Sun, Min Sun, and Hwann-Tzong Chen.
\newblock Direct voxel grid optimization: Super-fast convergence for radiance
  fields reconstruction.
\newblock In {\em Proceedings of the IEEE/CVF Conference on Computer Vision and
  Pattern Recognition}, pages 5459--5469, 2022.

\bibitem{sun2022improved}
Cheng Sun, Min Sun, and Hwann-Tzong Chen.
\newblock Improved direct voxel grid optimization for radiance fields
  reconstruction.
\newblock {\em arXiv preprint arXiv:2206.05085}, 2022.

\bibitem{sun2021neuralrecon}
Jiaming Sun, Yiming Xie, Linghao Chen, Xiaowei Zhou, and Hujun Bao.
\newblock Neuralrecon: Real-time coherent 3d reconstruction from monocular
  video.
\newblock In {\em Proceedings of the IEEE/CVF Conference on Computer Vision and
  Pattern Recognition}, pages 15598--15607, 2021.

\bibitem{wang2022go}
Jingwen Wang, Tymoteusz Bleja, and Lourdes Agapito.
\newblock Go-surf: Neural feature grid optimization for fast, high-fidelity
  rgb-d surface reconstruction.
\newblock {\em arXiv preprint arXiv:2206.14735}, 2022.

\bibitem{wang2021neus}
Peng Wang, Lingjie Liu, Yuan Liu, Christian Theobalt, Taku Komura, and Wenping
  Wang.
\newblock Neus: Learning neural implicit surfaces by volume rendering for
  multi-view reconstruction.
\newblock {\em arXiv preprint arXiv:2106.10689}, 2021.

\bibitem{weder2020routedfusion}
Silvan Weder, Johannes Schonberger, Marc Pollefeys, and Martin~R Oswald.
\newblock Routedfusion: Learning real-time depth map fusion.
\newblock In {\em Proceedings of the IEEE/CVF Conference on Computer Vision and
  Pattern Recognition}, pages 4887--4897, 2020.

\bibitem{yan2017distinguishing}
Qingan Yan, Long Yang, Ling Zhang, and Chunxia Xiao.
\newblock Distinguishing the indistinguishable: Exploring structural
  ambiguities via geodesic context.
\newblock In {\em Proceedings of the IEEE conference on computer vision and
  pattern recognition}, pages 3836--3844, 2017.

\bibitem{yariv2021volume}
Lior Yariv, Jiatao Gu, Yoni Kasten, and Yaron Lipman.
\newblock Volume rendering of neural implicit surfaces.
\newblock {\em Advances in Neural Information Processing Systems},
  34:4805--4815, 2021.

\bibitem{yariv2020multiview}
Lior Yariv, Yoni Kasten, Dror Moran, Meirav Galun, Matan Atzmon, Basri Ronen,
  and Yaron Lipman.
\newblock Multiview neural surface reconstruction by disentangling geometry and
  appearance.
\newblock {\em Advances in Neural Information Processing Systems},
  33:2492--2502, 2020.

\bibitem{yu2021plenoctrees}
Alex Yu, Ruilong Li, Matthew Tancik, Hao Li, Ren Ng, and Angjoo Kanazawa.
\newblock Plenoctrees for real-time rendering of neural radiance fields.
\newblock In {\em Proceedings of the IEEE/CVF International Conference on
  Computer Vision}, pages 5752--5761, 2021.

\bibitem{zhu2023nicer}
Zihan Zhu, Songyou Peng, Viktor Larsson, Zhaopeng Cui, Martin~R Oswald, Andreas
  Geiger, and Marc Pollefeys.
\newblock Nicer-slam: Neural implicit scene encoding for rgb slam.
\newblock {\em arXiv preprint arXiv:2302.03594}, 2023.

\bibitem{zhu2022nice}
Zihan Zhu, Songyou Peng, Viktor Larsson, Weiwei Xu, Hujun Bao, Zhaopeng Cui,
  Martin~R Oswald, and Marc Pollefeys.
\newblock Nice-slam: Neural implicit scalable encoding for slam.
\newblock In {\em Proceedings of the IEEE/CVF Conference on Computer Vision and
  Pattern Recognition}, pages 12786--12796, 2022.

\end{thebibliography}
